# Symptom-based Machine Learning Models for the Early Detection of COVID-19: A Narrative Review

Moyosolu Akinloye
*Department of Computing and Informatics*
*Bournemouth University*
Bournemouth, England, United Kingdom
s5550338@bournemouth.ac.uk

*Abstract*—Despite the widespread testing protocols for COVID-19, there are still significant challenges in early detection of the disease, which is crucial for preventing its spread and optimizing patient outcomes. Owing to the limited testing capacity in resource-strapped settings and the limitations of the available traditional methods of testing, it has been established that a fast and efficient strategy is important to fully stop the virus. Machine learning models can analyze large datasets, incorporating patient-reported symptoms, clinical data, and medical imaging. Symptom-based detection methods have been developed to predict COVID-19, and they have shown promising results. In this paper, we provide an overview of the landscape of symptoms-only machine learning models for predicting COVID-19, including their performance and limitations. The review will also examine the performance of symptom-based models when compared to image-based models. Because different studies used varying datasets, methodologies, and performance metrics. Selecting the model that performs best relies on the context and objectives of the research. However, based on the results, we observed that ensemble classifier performed exceptionally well in predicting the occurrence of COVID-19 based on patient symptoms with the highest overall accuracy of 97.88%. Gradient Boosting Algorithm achieved an AUC (Area Under the Curve) of 0.90 and identified key features contributing to the decision-making process. Image-based models, as observed in the analyzed studies, have consistently demonstrated higher accuracy than symptom-based models, often reaching impressive levels ranging from 96.09% to as high as 99%.

## I. INTRODUCTION

COVID-19, a virus caused by SARS-CoV-2, has had a global impact, affecting over 175 million individuals worldwide between December 2019 and June 2021[1]. Healthcare infrastructures in many countries, like that of the NHS (National Health Service) in the United Kingdom, have been under intense pressure due to the rising number of critical cases [2]. To alleviate this burden, prompt detection of COVID-19 cases is essential [1]. This paper provides an overview of the landscape of symptoms-only machine learning models for predicting COVID-19, encompassing their performance, limitations, and prospects.

Widespread testing for COVID-19 is crucial as it helps us understand the pattern of transmission of the virus, and if interventions are working [3]. To do this, the Percentage of positive COVID-19 tests, also known as Percent Positive (PP), out of all the tests done in a brief time should be assessed. A high number of PP means the virus is spreading a lot among people, which also means that lots of tests need to be carried out. Alternatively, a low PP is an indicator that the virus is spreading less [4]. The WHO (World Health Organization) recommends targeting a PP below 5% as a measure to manage and curb the spread of the pandemic. When a lot of tests are carried out with lower PP, the efforts of interventions are amplified, and this slows down the spread of the pandemic. Testing is therefore important in fighting the pandemic as it is the first step to detecting and diagnosing the virus.

According to Our World in Data, Mexico has performed just three tests per 100,000 people (about the seating capacity of the Los Angeles Memorial Coliseum) daily since the pandemic started. This is the second lowest testing rate among other countries that have been most impacted by the pandemic. This reported data is the same when compared with other developing and some developed countries [5] and is assumed to be due to limited testing capacity as countries were unprepared and their health systems not fully equipped to fight the pandemic. To optimize limited testing capacity, an approach involves carefully checking individuals to determine potential virus exposure. Instead of testing everyone, only those who are suspected of having the virus are tested. This checking involves talking to patients to see if they might have been exposed to the virus [6]. When governments apply suitable measures, they can lower local transmission rates, gaining control over the pandemic [7].

Tests like real-time reverse transcription-polymerase chain reaction (RT-PCR) have been important in helping to identify and manage the pandemic. It is the gold standard on which diagnosis of COVID-19 relied on at the early stage of the pandemic. There is a limitation of testing accessibility and the presence of asymptomatic and pre-symptomatic carriers. Additionally, conventional PCR tests may not detect the virus in the upper respiratory tract during the initial week following the onset of symptoms. This is because the presence of the virus in this area diminishes during this initial week after symptoms appear [8]. X-rays and Computed Tomography (CT) scans have also been employed in the battle against COVID-19, but they are expensive to be deployed for this purpose and difficult to scale up so that its benefits can be available to everyone, especially in places with limited resources.

Due to the lack of testing capacity in limited resource settings like Mexico that have been mentioned, and the limitation of the available traditional methods of testing, it has been established that a fast and effective strategy is important to fully stop the virus.

In response to the pandemic, the scientific community has focused on creating solutions, with a significant emphasis on applying machine learning and artificial intelligence (AI) to COVID-19-related issues [9]. Machine learning models can analyze large datasets, incorporating patient-reported symptoms, clinical data, and medical imaging. Symptom-based detection methods have been developed to predict COVID-19, and they have shown promising results [10]. However, some studies have suggested that symptoms alone are not sufficient to find the effectiveness of machine learning models, thereby leaving a question of whether COVID-19 could be diagnosed early enough to prevent complications and spread of the disease, especially in settings where resources are scarce.

In context of the evolving pandemic, characterized by the introduction of vaccines, the emergence of new variants, and changing public health responses, understanding the role and potential of the symptom-based machine learning models remains essential. They not only facilitate early detection but also contribute to our broader comprehension of the virus dynamics. By providing insight into the efficacy of symptom-based prediction and its place within the broader public health toolkit, this review aims to inform and guide ongoing efforts in pandemic response and preparedness. The research questions guiding this investigation encompass understanding how effective symptom-based machine learning models are for early detection of COVID-19, what the strengths and weaknesses of existing symptom-based models are in predicting COVID-19, and how symptom-based models compare with image-based models in terms of accuracy and performance.

Following this introduction, Section II delves deeper into the identified challenges in early disease detection and the potential solutions proposed by symptom-based machine learning models. Section III provides an in-depth analysis of studies conducted in this domain, categorizing them into symptom-based and image-based models, discussing their methodologies, findings, and performance metrics. Section IV offers a critical review of the literature, synthesizing the key insights derived from these studies and their implications for COVID-19 detection. Finally, Section V consolidates these findings into a conclusive perspective, offering a comparative analysis between symptom-based and image-based models and highlighting their respective strengths and limitations, provides recommendations and points to future works.

## II. THE PROBLEM

Despite the widespread testing protocols for COVID-19, there are still significant challenges in early detection of the disease, which is crucial for preventing its spread and optimizing patient outcomes. Traditional testing methods, such as PCR tests can be time consuming, expensive and resource intensive. Moreover, they may not always be readily available, especially in remote settings.

This study aims to explore the potential of symptom-based machine learning models as a solution to this problem. It aims to review the effectiveness of these models in achieving early detection of the disease. These models could potentially analyze a combination of symptoms to predict the likelihood of a person having the disease, thereby facilitating early detection and intervention. However, the effectiveness and accuracy of these models have yet to be thoroughly reviewed and understood. Hence this study seeks to fill this knowledge gap.

## III. RELATED WORKS

Existing research has been organized and categorized to facilitate understanding. Application-based taxonomy, which grouped COVID-19-related research into three major categories, was proposed. These categories of Covid-19 research are ongoing with dozens of new articles published weekly as researchers continue to improve on the existing results [9]. One of the categories primarily revolves around finding and diagnosing the disease, which is sub-categorized into two main groups: one focuses on using symptoms and clinical assessments to find COVID-19 cases, while the other uses images of patients' lungs from X-rays and CT scans to detect the presence of the disease.

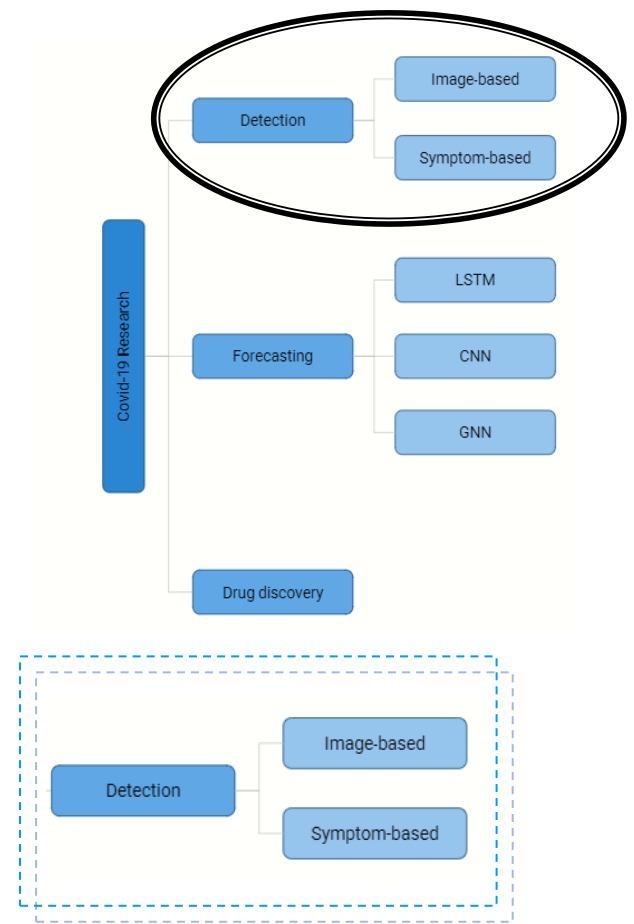

### A. Symptom-based Models

The authors in [12] explored the use of hybrid and ensemble models to assist healthcare units in identifying COVID-19 cases based on symptoms alone, before conducting physical tests. The goal is to provide early predictions to help healthcare providers prepare necessary precautions and prevent unmanaged cases from causing future problems. The study initially employed various traditional classification models like Logistic Regression, K-nearest Neighbors (KNN) classifier, Entropy-based decision tree classifier, Random Forest classifier, and Gradient Boosting classifier to analyze symptom data. The results indicated that Gradient Boosting and Random Forest classifiers stood out as the most promising choices due to their high accuracy and acceptable type II errors. Subsequently, the study implemented a hybrid approach involving clustering followed by classification. This hybrid model outperformed all the traditional models, achieving an improved accuracy rate of 87%. Finally, the Max-Voting ensemble model, which combined Random Forest and Gradient Boosting algorithms, further enhanced accuracy to approximately 90%, making it the preferred approach for the given dataset.

The author in [13] compared five machine learning algorithms- support vector machines (SVM), random forest (RF), multilayer perceptron (MLP), gradient boosting decision trees (GBDT), and logistic regression - to diagnose Covid-19 patients using the symptoms that they exhibited upon their clinic admission. The study found that the SVM classifier achieved the highest accuracy, reaching an AUC (Area Under the Curve) of 0.85.

Another comprehensive study [14] utilized a dataset which encompasses 8 essential clinical characteristics. The researchers attained an impressive AUC of 0.90 by employing the gradient boosting algorithm. The model highlighted key characteristics that strongly impacted the decision-making process. It identified symptoms like fever and cough as critical indicators for predicting disease contraction. Also, being close to someone confirmed with COVID-19 emerged as a vital feature, highlighting the high transmissibility of the disease

In a recent study by [15], the authors employed an ensemble classifier to construct a model based on patient symptoms. The proposed model accurately predicted disease occurrence at 97.88% overall accuracy.

In another recent study by [16], the authors compared the performance of several machine learning models to find Covid-19 based on the patient's symptoms. The study found that Naive Bayes and decision tree classifiers performed best with an accuracy of 93.70%, followed by SVM, k-nearest neighbors, and logistic regression, that achieved an accuracy of 93.60%, 93.50%, and 92.80% respectively.

The authors of [11] used logistic regression on a dataset comprising only 64 confirmed positive cases. To gauge the effectiveness of the classifier, they employed the area under the curve for the receiver characteristic (AUC ROC) as an evaluation metric. For SARS-CoV-2 and the remaining common respiratory viruses, the predictability based solely on presenting symptoms was not high. The average AUROCs for these viruses were below 0.70, suggesting that relying solely on the symptoms a patient shows might not be enough to accurately determine if they will test positive for these viruses. These findings are consistent with earlier studies that also concluded symptoms alone were not adequate for accurately diagnosing influenza or differentiating it from other illnesses like influenza. The study's limitations include the absence of information on the duration of reported symptoms in the models. Additionally, the small number of SARS-CoV-2 positive cases in the dataset and variations in testing protocols within health systems were noted as limitations. The evolving understanding of COVID-19 symptoms and the inclusion of emerging symptoms like gastrointestinal and dermatologic symptoms were highlighted as factors impacting the accuracy of symptom-based models. The study acknowledges that clinicians' knowledge of COVID-19 symptoms is evolving, and ongoing data collection is essential to assess the impact of improved symptom characterization and additional data on the performance of models in identifying SARS-CoV-2 infections.

In [17], the authors used a hybrid method, merging wearable sensor data and self-reported symptoms for COVID-19 prediction. Instead of training a classifier, they conducted a statistical analysis and reported a sensitivity of 0.72 and a specificity of 0.73, respectively.

Other research was found to have employed voice signals and symptoms. Audio recordings of people coughing (both infected and not infected with COVID-19), were used to train Support Vector Machines (SVM); Their finding showed a sensitivity of 0.68 and a specificity of 0.82 [18].

Another example is the work presented in [19], where the researchers revealed the variations in symptoms among infected patients. They categorized 1653 infected participants into six groups using clustering. The clusters were used to predict the need for respiratory support rather than for infection detection.

In [20] The research aimed to create a tool combining a clinical prediction rule and a point-of-care test for diagnosing COVID-19 in symptomatic outpatients. Before conducting SARS-CoV-2 PCR testing, they used a standardized clinical questionnaire, gathering data unbeknownst to physicians. By combining various symptoms into 326 patterns and employing logistic regression, they identified indicators of COVID-19. This resulted in a scoring system with different likelihood ratios for different scores. The development phase involved 120 positive and 120 negative COVID-19 cases, validated with 40 additional cases. Notable predictors of COVID-19 included the loss of taste or smell and fever with cough, while wheezing and chest tightness pointed towards non-COVID-19 respiratory viral infections. The model's accuracy measured by AUC was 0.736.

Some studies did not report specific metrics, such as sensitivity, specificity, or AUC. Accuracy is provided where available. The choice of metrics varies depending on the study's goals and methodologies.

*B. Image-Based Models*

In COVID-19 detection, two primary approaches are used: feature extraction from images and transfer learning with pretrained deep-learning models. In the feature extraction approach, authors manually select or use machine learning algorithms like Iso-map or Principal Component Analysis for feature extraction [21]. One study achieved an accuracy of 96.09% and 98.09% using Fractional Multichannel Exponent Moments (FrMEMs) and special optimization techniques [21]. Another study applied U-Net models for lung and COVID-19 lesion segmentation and achieved an AUC of 0.927 [22]. In the transfer learning approach, pretrained models are used either to extract features or to train the final layers on new data [23]. This method tackles the issue of limited training data within the COVID-19 datasets. Notably, it allows using established pretrained models. One study reported a sensitivity of

98.66%, a specificity of 96.46% and a top accuracy of 96.78% [23]. Other research explored the combination of various pretrained models and classifiers, achieving 99% accuracy in one case [24]. CNN-based (Convolutional Neural Network) ensemble methods demonstrated robust results [25]. COVID-19 image-based detection methods have shown impressive performance, with accuracy rates of up to 99% [21]. In summary, these studies employ different strategies for COVID-19 detection using image data, achieving high accuracy rates but facing challenges in deployment. Rigorous testing and cost considerations are crucial for practical implementation.

*C. Review of the Literatures*

From the studies, it is challenging to definitively conclude which author's model performed the best because different studies used varying datasets, methodologies, and performance metrics. The choice of the best-performing model also depends on the specific context and goals of the study. However, we can make some observations based on the observed results:

Study [15] reported the highest overall accuracy of 97.88%, which is an impressive result. This suggests that the ensemble classifier used in this study performed exceptionally well in predicting the occurrence of COVID-19 based on patient symptoms. Study [14] achieved an AUC of 0.90 using the gradient boosting algorithm and identified key features contributing to the model's decision-making process. This indicates strong predictive capabilities for COVID-19 diagnosis based on symptoms. Study [16] reported high accuracy for Naive Bayes and decision tree classifiers (93.70%) and high accuracy for SVM, KNN, and Logistic Regression (around 93% or higher). These models performed well in diagnosing COVID-19 based on symptoms. Study [13] showed that the SVM classifier had the highest accuracy with an AUC of 0.85. While not the highest accuracy reported, it performed well in diagnosing COVID-19 based on symptoms. Study [12] used hybrid and ensemble models, achieving an accuracy of approximately 90%. The Max-Voting ensemble model was preferred, indicating its strong predictive power.

Different models and approaches have shown promising results in predicting COVID-19 based on symptoms. Study [15] reported the highest overall accuracy, but other studies, such as [14], [16], and [13], also achieved strong results. The choice of the best-performing model is context-dependent and may require further evaluation in a specific clinical setting. It is also important to consider the trade-offs between sensitivity, specificity, and other factors when selecting a model for practical use in healthcare settings.

When we place the symptom-based model and the image-based model side-by-side, the latter achieved prominent levels of performance, with other studies reporting impressive results.

Study [24] achieved an accuracy of 99% using a combination of various pretrained models and classifiers. This represents the highest reported accuracy among the studies mentioned. Study [21] reported accuracy rates of 96.09% and 98.09% using FrMEMs and special optimization techniques for feature extraction. While not as high as the accuracy in [24], it is still noteworthy. Study [23] reported a top accuracy of 96.78% with high sensitivity (98.66%) and specificity (96.46%) by using transfer learning and pretrained models. This study achieved a balanced performance across multiple metrics. Study [22] used U-Net models for lung and COVID-19 lesion segmentation and achieved an AUC of 0.927. While it did not report accuracy, it demonstrated impressive performance in segmentation tasks. Study [23] employed CNN-based ensemble methods that demonstrated robust results. While specific accuracy figures are not provided, the robustness of the approach is noteworthy.

As in the symptoms-based models, the choice of the "best" image-based model depends on the specific requirements and objectives of the application. If high accuracy is the primary criterion, then the model in study [24] with 99% accuracy stands out. However, other factors like sensitivity, specificity, and practical considerations may also influence the selection of the best model.

In summary, study [24] reported the highest accuracy, but the selection of the best image-based model should consider the specific needs of the application and may require trade-offs between different performance metrics.

IV. DISCUSSIONS

After the Image-based models, as seen in studies like [21], have consistently demonstrated impressive accuracy rates, with results ranging from 96.09% to as high as 99%. These models leverage feature extraction from medical images, such as X-rays and CT scans, and often employ advanced deep learning techniques and transfer learning [23]. For example, study [22] applied U-Net models for lung and COVID-19 lesion segmentation and achieved an AUC of 0.927. Image-based models offer the advantage of analyzing direct visual information obtained from medical imaging, enabling precise lesion identification. However, these models face deployment challenges, including the need for rigorous testing and considerations regarding the cost and resource-intensive nature of medical imaging infrastructure.

In contrast, symptom-based models, as exemplified in [15] and [16], rely on self-reported symptoms, clinical questionnaires, or audio data (such as cough sounds) for COVID-19 detection. While some symptom-based models have reported high accuracy, typically in the 90s, they exhibit lower accuracy compared to image-based models. These models frequently use traditional machine learning methods and clinical prediction rules [20]. Challenges associated with symptom-based models include the evolving understanding of COVID-19 symptoms and the impact of data limitations, including variations in testing protocols. However, they offer accessibility and cost-effectiveness, making them practical in scenarios where medical imaging resources are limited or impractical.

The choice between image-based and symptom-based models should consider the specific context and objectives of the application. In some scenarios, image-based models, as demonstrated in study [24], may provide the most accurate and direct detection of COVID-19. Conversely, symptom-based models, as in [16], can serve as valuable tools, especially when medical imaging resources are not readily available, while also offering practicality in terms of cost and accessibility. Furthermore, a comprehensive strategy may involve the combination of both approaches, as the strengths of each can complement one another and offer improved accuracy and diagnostic capabilities in the ongoing fight against COVID-19.

## V. CONCLUSIONS, RECOMMENDATIONS, AND FUTURE WORK

### A. Conclusion

The comparison of symptom-based and image-based models for COVID-19 detection offers valuable insights and implications. Image-based models, as observed in the analyzed studies, have consistently demonstrated higher accuracy, often reaching impressive levels ranging from 96.09% to as high as 99% [21]. This is particularly significant when reliable medical imaging data, such as X-rays or CT scans, is readily available. The utilization of advanced techniques like deep learning and transfer learning has proven effective in enhancing the performance of image-based models, as seen in the study that applied U-Net models for lung and COVID-19 lesion segmentation, achieving an AUC of 0.927 [22]. These models excel in precise lesion identification and can play a pivotal role in the diagnosis of COVID-19. However, they do come with challenges, including the need for rigorous testing and considerations related to the cost and resource-intensive nature of medical imaging infrastructure.

On the other hand, symptom-based models, while achieving lower accuracy, offer advantages in terms of accessibility and cost-effectiveness [15] [16]. They rely on self-reported symptoms, clinical questionnaires, or audio data, such as cough sounds. As demonstrated in the analyzed studies, certain symptom-based models have achieved high accuracy, typically in the 90s [16]. They frequently employ traditional machine learning methods and clinical prediction rules, and they can be particularly valuable when medical imaging resources are limited or impractical, making them more widely applicable.

### B. Recommendation

Considering the strengths of both approaches, a recommended strategy would involve a thoughtful selection of the detection method based on the specific context and objectives. In cases where reliable medical imaging data is available and resources permit, image-based models, as demonstrated in a study achieving a top accuracy of 99% [24], can serve as a powerful tool for accurate COVID-19 detection. On the other hand, when resources or imaging infrastructure are constrained, symptom-based models, such as those reported in a study with a high accuracy rate of 97.88% [16], become a valuable choice. However, a truly comprehensive and potentially more accurate approach could involve combining both symptom-based and image-based models, leveraging the strengths of each to enhance overall diagnostic accuracy. Collaboration between medical professionals, data scientists, and AI researchers is crucial for the development of models that are not only accurate but also clinically relevant, scalable, and deployable.

### C. Future Work

In terms of future works, ongoing research and refinement of both image-based and symptom-based models are essential to further enhance their performance. Development of hybrid models that integrate multiple data sources, including medical images and clinical symptoms, holds promise for achieving even more accurate and robust COVID-19 detection. The impact of emerging symptoms, variations in testing protocols, and the evolving understanding of COVID-19 should be further explored to improve symptom-based models [16]. Deployment studies that assess the real-world applicability of these models and their integration into healthcare settings are critical for practical implementation. Collaborative efforts between researchers, healthcare providers, and policymakers are vital to address the unique challenges and opportunities presented by different detection approaches and to ensure the most effective tools are deployed in the battle against COVID-19. In conclusion, the selection and combination of symptom-based and image-based models should be guided by the specific context and available data, with the overarching goal of improving the accuracy of COVID-19 detection to enhance disease management and public health efforts.

## REFERENCES

[1] L. S. Canas, "Early detection of COVID-19 in the UK using self-reported symptoms: a large-scale, prospective, epidemiological surveillance study", *The Lancet Digital Health*, vol. 3, no. 9, pp. e587–e598, 2021, doi: 10.1016/s2589-7500(21)00131-x.

[2] M. Al-Hussaini, A. H. Mansour, H. Morreim, M. H. Zawati, and T. A. Arawi, Bioethics amidst the COVID-19 pandemic. Frontiers Media SA, 2022.

[3] O. M. Araz, A. Ramirez-Nafarrate, M. Jehnand F. A. Wilson, "The importance of widespread testing for COVID-19 pandemic: systems thinking for drive-through testing sites", *Health Systems*, vol. 9, no. 2, pp. 119–123, 2020, doi: 10.1080/20476965.2020.1758000.

[4] "COVID-19 testing: Understanding the 'Percent positive,'" Johns Hopkins Bloomberg School of Public Health, Aug. 04, 2021. [Online]. Available: https://publichealth.jhu.edu/2020/covid-19-testing-understanding-the-percent-positive#:~:text=The%20percent%20positive%20will%20be,haven't%20been%20tested%20yetJ. Hasell, "A cross-country database of COVID-19 testing", *Scientific Data*, vol. 7, no. 1, 2020, doi: 10.1038/s41597-020-00688-8.

[5] J. Hasell, "A cross-country database of COVID-19 testing", Scientific Data, vol. 7, no. 1, 2020, doi: 10.1038/s41597-020-00688-8.

[6] R. Martinez-Velazquez, D. P. Tobón V., A. Sanchez, A. El Saddikand E. Petriu, "A Machine Learning Approach as an Aid for Early COVID-19 Detection", *Sensors*, vol. 21, no. 12, p. 4202, 2021, doi: 10.3390/s21124202.

[7] J. Peto, "Covid-19 mass testing facilities could end the epidemic rapidly", *BMJ*, p. m1163, 2020, doi: 10.1136/bmj.m1163.

[8] M. J. Binnicker, "Challenges and controversies to testing for COVID-19," Journal of Clinical Microbiology, vol. 58, no. 11, Oct. 2020, doi: 10.1128/jcm.01695-20.

[9] F. Kamalov, A. K. Cherukuriand F. Thabtah, "Machine learning applications to Covid-19: a state-of-the-art survey", 2022. doi: 10.1109/aset53988.2022.9734959.

[10] M. A. Callejon-Leblic, "Loss of Smell and Taste Can Accurately Predict COVID-19 Infection: A Machine-Learning Approach", *Journal of Clinical Medicine*, vol. 10, no. 4, p. 570, 2021, doi: 10.3390/jcm10040570.

[11] A. Callahan, "Estimating the efficacy of symptom-based screening for COVID-19", *npj Digital Medicine*, vol. 3, no. 1, 2020, doi: 10.1038/s41746-020-0300-0.

[12] C. Koushik, R. Bhattacharjee and C. S. Hemalatha, "Symptoms based Early Clinical Diagnosis of COVID-19 Cases using Hybrid and Ensemble Machine Learning Techniques", 2021. doi: 10.1109/icccsp52374.2021.9465494.

[13] L. Wynants, "Prediction models for diagnosis and prognosis of covid-19: systematic review and critical appraisal", *BMJ*, p. m1328, 2020, doi: 10.1136/bmj.m1328.

[14] Y. Zoabi and N. Shomron, "COVID-19 diagnosis prediction by symptoms of tested individuals: a machine learning approach", 2020. doi: 10.1101/2020.05.07.20093948.

[15] A. Kumari and A. K. Mehta, "Effective Prediction of COVID-19 Using Supervised Machine Learning with Ensemble Modeling," Algorithms for Intelligent Systems, pp. 537–547, 2022, doi: https://doi.org/10.1007/978-981-16-5747-4_45.

[16] M. Malik, "Determination of COVID-19 Patients Using Machine Learning Algorithms", *Intelligent Automation & Soft Computing*, vol. 31, no. 1, pp. 207–222, 2022, doi: 10.32604/iasc.2022.018753.

[17] G. Quer, "Wearable sensor data and self-reported symptoms for COVID-19 detection", *Nature Medicine*, vol. 27, no. 1, pp. 73–77, 2021, doi: 10.1038/s41591-020-1123-x.

[18] J. Han, "Exploring Automatic COVID-19 Diagnosis via Voice and Symptoms from Crowdsourced Data", 2021. doi: 10.1109/icassp39728.2021.9414576.

[19] C. H. Sudre et al., "Symptom clusters in COVID-19: A potential clinical prediction tool from the COVID Symptom Study app," Science Advances, vol. 7, no. 12, Mar. 2021, doi: 10.1126/sciadv.abd4177.

[20] D. S. Smith, E. A. Richey and W. L. Brunetto, "A Symptom-Based Rule for Diagnosis of COVID-19", *SN Comprehensive Clinical Medicine*, vol. 2, no. 11, pp. 1947–1954, 2020, doi: 10.1007/s42399-020-00603-7.

[21] M. A. Elaziz, K. M. Hosny, A. Salah, M. M. Darwish, S. Lu and A. T. Sahlol, "New machine learning method for image-based diagnosis of COVID-19", *PLOS ONE*, vol. 15, no. 6, p. e0235187, 2020, doi: 10.1371/journal.pone.0235187.

[22] W. Cai, "CT Quantification and Machine-learning Models for Assessment of Disease Severity and Prognosis of COVID-19 Patients", *Academic Radiology*, vol. 27, no. 12, pp. 1665–1678, 2020, doi: 10.1016/j.acra.2020.09.004.

[23] I. D. Apostolopoulos and T. A. Mpesiana, "Covid-19: automatic detection from X-ray images utilizing transfer learning with convolutional neural networks", *Physical and Engineering Sciences in Medicine*, vol. 43, no. 2, pp. 635–640, 2020, doi: 10.1007/s13246-020-00865-4.

[24] S. H. Kassania, P. H. Kassanib, M. J. Wesolowskic, K. A. Schneidera, and R. Detersa, "Automatic Detection of coronavirus disease (COVID-19) in x-ray and CT images: A Machine learning based approach," Biocybernetics and Biomedical Engineering, vol. 41, no. 3, pp. 867–879, Jul. 2021, doi: 10.1016/j.bbe.2021.05.013.

[25] P. Saha, M. S. Sadi, and Md. M. Islam, "EMCNet: Automated COVID-19 diagnosis from X-ray images using convolutional neural network and ensemble of machine learning classifiers," Informatics in Medicine Unlocked, vol. 22, p. 100505, Jan. 2021, doi: 10.1016/j.imu.2020.100505.